\documentclass[acmtog]{acmart}
\usepackage{natbib}
\usepackage{array}
\usepackage{colortbl}
\usepackage{xcolor}

\AtBeginDocument{%
  \providecommand\BibTeX{{%
    \normalfont B\kern-0.5em{\scshape i\kern-0.25em b}\kern-0.8em\TeX}}}

\copyrightyear{2023}
\acmYear{2023}
\acmDOI{}

\begin{document}

\title{Blar-SQL: Faster, Stronger, Smaller NL2SQL}

\author{José Manuel Domínguez}
\email{jose@getblar.com}
\affiliation{%
  \institution{Blar Spa}
  \streetaddress{}
  \city{Santiago}
  \state{}
  \country{Chile}
  \postcode{}
}

\author{Benjamín Errazuriz}
\email{benjamin@getblar.com}
\affiliation{%
  \institution{Blar Spa}
  \streetaddress{}
  \city{Santiago}
  \state{}
  \country{Chile}
  \postcode{}
}

\author{Patricio Daher}
\email{patricio@getblar.com}
\affiliation{%
  \institution{Blar Spa}
  \streetaddress{}
  \city{Santiago}
  \state{}
  \country{Chile}
  \postcode{}
}

\renewcommand{\shortauthors}{Dominguez and Errazuriz}

\begin{abstract}
    Large Language Models (LLMs) have gained considerable notoriety in the field of natural language to SQL tasks (NL2SQL). In this study, we show how task decomposition can greatly benefit LLMs in database understanding and query generation in order to answer human questions with an SQL query.
    
    We fined-tuned open source models, specifically Llama-2 and Code Llama, by combining 2 different models each designated to focus on one of two tasks in order to leverage each model's core competency to further increase the accuracy of the final SQL query.
    
    We propose a new framework to divide the schema into chunks in order to fit more information into a limited context. Our results are comparable with those obtained by GPT-4 at the same time being \textasciitilde 135 times smaller, 90 times faster and more than 100 times cheaper than GPT-4.
\end{abstract}


\keywords{NL2SQL, LLM, AI, Prompt Decomposition}

\received{24 October 2023}

\maketitle

\section{Introduction}
With recent advances in LLMs such as the Open AI's GPT-4 and Meta's Llama-2, the field of translating human language questions into SQL queries, commonly referred to as NL2SQL, has witnessed significant advancements.  These state-of-the-art language models have opened up exciting opportunities for improving the accuracy and efficiency of NL2SQL systems, enabling the end user to interact with databases without having a deep understanding of SQL language or the database's schema. 

Currently, most of the SOTA approaches use LLMs to achieve this goal, particularly GPT-4, which has gained most of the attention as being the largest and most capable model as of this paper's writing. In this paper, we explore how open-source models such as Llama-2 and CodeLlama when fine-tuned can achieve similar or even better results than the ones achieved by GPT-4. Moreover, we explore how dividing complex tasks into smaller steps can significantly benefit these models. We build on DIN-SQL's approach \cite{pourreza2023din} and fine-tune two custom models, each specialized in completing one sub-task achieving results comparable with GPT-4 on the BIRD-SQL data set. 

Another challenge frequently encountered in the text-to-SQL domain is managing context. Large data schemas often exceed the 32K token context limit of the GPT-4 model. To address this, we have introduced a novel method for selecting the necessary schema link to answer a given query. We partition the schema into several prompts, each tailored to fit within the maximum context length. This approach enables models with a context limit of 4K tokens or less to efficiently process databases with extensive schemas and descriptions.

The contributions of this study are summarized:
\begin{enumerate}
    \item We divide the task of generating SQL queries into 2 steps, the first is choosing the appropriate columns based on the database schema, column descriptions and other external knowledge that may be required.
    \item A new framework of sub-dividing very large schemas and descriptions into chunks in order to better manage prompt context and include the greatest amount of information possible.
    \item A new idea of using 2 different models trained with different purposes in order to leverage their knowledge and create better results.
\end{enumerate}

\section{Related Work}

Decomposing complex text-to-SQL tasks has shown great potential, improving the performance of few shots LLMs on the Spider dataset by over 10\% \cite{pourreza2023din}. The main idea behind this is that LLMs respond better when only one task is asked, to achieve this you break the problem into simple pieces and solve each part using prompting, trained models, or symbolic functions \cite{khot2022decomposed}.

SDSQL model introduces the notion of capturing the interactions between the question and the data schema. It uses a series of dependency labels that link question tokens with schema tokens (\textit{W-Col}, \textit{S-Agg}, \textit{W-Op}).

There are several datasets designed for text-to-SQL research, such as Spider \cite{yu2018spider} and WikiSQL \cite{zhong2017seq2sql}. These datasets not only provide a benchmark for comparing various approaches, but they also supply the data necessary for fine-tuning models and achieving significant improvements. In this realm, the BIRD dataset stands out. It encompasses a 33.4GB database emphasizing real-world applications. This benchmark offers a detailed insight into how our frameworks would perform with actual databases, making it the most rigorous benchmark available \cite{li2023can}.

\section{Methodology}

\subsection{Error analysis}

In order to further understand the current capabilities of open-source LLMs we first evaluated the vanilla version of code-llama against the development dataset of both Spider and Bird. We later took a random subset of 60 data points and classified the errors in order to understand what should be improved. 

The most common error we found where schema linking, this occurred when the model changed the name of some columns or hallucinated columns and table-column associations. The second most common error was when the model failed to understand the content of the database, this error occurred when the model couldn't comprehend the column description and values. Figures 1 and 2 show examples of both errors.  
\begin{figure}[h]
  \centering
  \includegraphics[width=\linewidth]{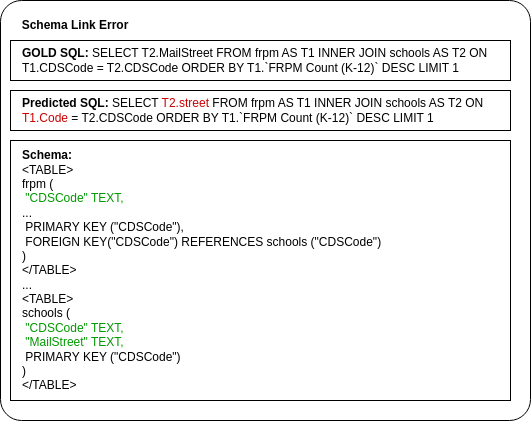}
  \caption{Example of a Schema Link Error}
  \smallskip  
  \parbox{\linewidth}{\small
    In this example the model got wrong the name of the columns 'CDSCODE' and 'MailStreet'
  }
\end{figure}

\begin{figure}[h]
  \centering
  \includegraphics[width=\linewidth]{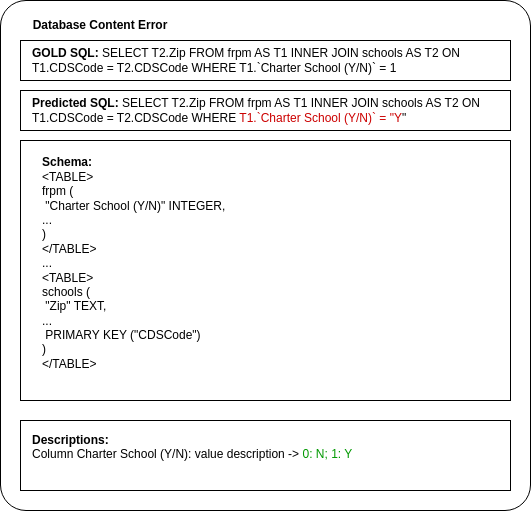}
  \caption{Example of a database content error}
  \parbox{\linewidth}{\small
    In this example the model assumed that the column 'Charter School (Y/N)' had values "Y" or "N", but in reality the values are 1 or 0
  }
\end{figure}

Our results coincide with the results obtained by the team that developed the Bird data set \cite{li2023can}. The two most common mistakes were "Wrong Schema Linking" (41.6\%) and "Misunderstanding Database Content" (40.8\%), both occurred because the LLM couldn't understand the database schema correctly or recall the correct structure of the database.

In order to further improve LLMs capability we divided the process into two steps: 1) inferring the schema links (Schema-Link model) and 2) Constructing the SQL using the previously generated schema links (SQL model). The overall architecture is shown in figure 3. The idea is to train both models specifically for their task because it has been proven that dividing a problem into smaller tasks generates better results when it comes to LLM inference \cite{khot2022decomposed}.

\begin{figure}[h]
  \centering
  \includegraphics[width=\linewidth]{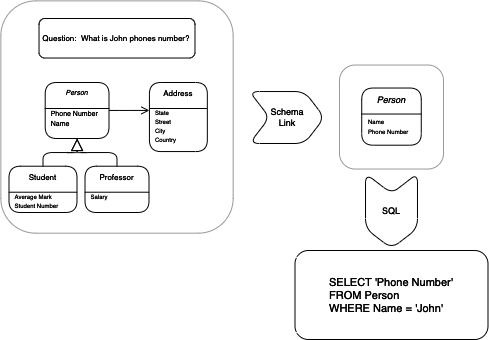}
  \caption{Prediction solution architecture}
  \parbox{\linewidth}{\small
    Here you can see in the first box the question and the database schema, which the Schema-Link model take and infer the tables and columns needed to answer the question. With the schema link the SQL model predicts the final SQL to answer the question.
  }
\end{figure}

\subsection{Making the data set}

As we mentioned before we used the Bird dataset because is the first benchmark focused on big databases with real applications. Thanks to the work done by the Bird team, you can access 95 databases with 12,751 questions with their respective answer \cite{li2023can}. 

In order to create the schema link dataset we reverse-engineered the SQL answers and the schema. This process consisted of extracting only the tables and columns used in the query answers as well as the foreign keys. Figure 4 shows the process of schema link generation. In order to achieve this we used a combination of SQL metadata capturing packages (\textit{sql-metadata} and \textit{sql-parse}) for Python combined with custom-made Regex. 

\begin{figure}[h]
  \centering
  \includegraphics[width=\linewidth]{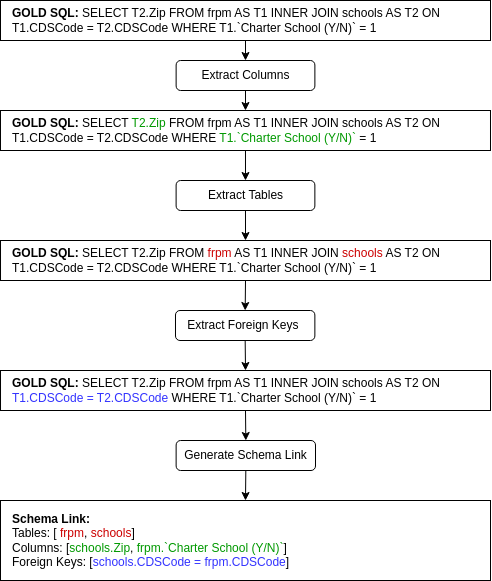}
  \caption{Schema Link Generation Process}
\end{figure}

\subsection{Hypothesis}

Our solution consists of two models working on two separate tasks, making the schema linking and generating the SQL query. The way that we train each model can impact how these two models interact with each other to come up with a solution. Because of this, we created two hypotheses:

\begin{enumerate}
    \item A non-trusting interaction: where the SQL model does not rely only on the Schema-Link output, it also has access to the original schema and column descriptions. The schema link are presented as a suggestion more than a fact. By ensuring that the SQL model does not solely rely on the output of the first, there's an added opportunity to correct potential misinterpretations of the schema.
    \item A trusting interaction where the SQL model only takes the output of the Schema-Link model to construct the query. This enables us to manage the context length as well as teach the model to focus purely on the construction of the SQL query and the logic behind it.
\end{enumerate}

Furthermore, we hypothesized that using a Llama-2 model for the schema linking part would be beneficial as it has a better world understanding \cite{roziere2023code} and using code-llama for the SQL part, as it has a better programming background \cite{touvron2023llama}. Unfortunately, due to a lack of time and resources we couldn't compare how a pure llama-2 and pure code-llama solution would perform, so we leave this as a future work. 

\subsection{Context Management}

One big limitation that open-source solutions such as llama-2, code-llama or others have, is the smaller context they have in comparison with Open AI's GPT models. Even GPT-4 32K context may fall short when the database is big enough or if it has detailed column descriptions.

We experimented with two approaches in order to tackle this problem. The first one consisted of reducing the prompt context by excluding certain information such as column descriptions if it exceeded the context. This traditional approach has the limitation of the volume of context that we can provide our models to make a prediction. We call this the non-descriptive approach.

The second approach is a more novel one, in which we propose a new method of dividing the schema linking step according to the maximum context length available. 

First, we started by creating the template of the prompt which consisted of four parts. The question, the database schema, column descriptions and finally a hint or external knowledge required in order to answer the question. 

Then, we iterated over each table's columns and descriptions, adding one table at a time until the prompt could not further fit another table. The question and external knowledge were fixed in order to always be included. We then divided each question using this method in one or more prompts.

Finally, we extracted the schema links from each segmented prompt, producing a list of schema link predictions. When concatenated, this list provided the essential information required to answer the question. We call this the schema chunking approach and Figure 5 explains the whole process.

\begin{figure}[h]
  \centering
  \includegraphics[width=\linewidth]{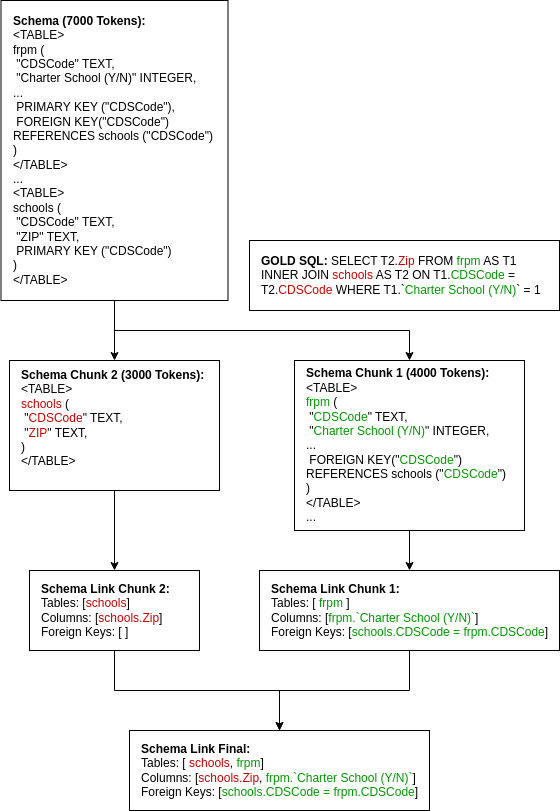}
  \caption{Schema chunking process}
\end{figure}

\subsection{Training process}

To train the models we used Google Colab's A100 40GB GPU. To achieve this we used 4bit model quantization as well as QLoRA for fine-tuning. The training itself used a combination of the packages available in huggingface such as \textit{transformers}, \textit{peft} and \textit{trl}. The training process usually lasted \textasciitilde10hrs costing about \$20 USD per model.

We did supervised fine-tuning only updating weights from the response (Schema links and SQL queries) similar to the approach taken by Dail-SQL \cite{gao2023text}. For the dataset, we used BIRD's Training set consisting of 8,952 examples, the dataset was subsequently partitioned randomly, resulting in 7,609 examples left for training, while 1,343 were set aside for validation purposes. 

We conducted the training process for a maximum of two epochs, evaluating the model every 500 steps and selecting the best models based on the validation set.

\subsubsection{Schema Link models}
\begin{itemize}
    \item Non-descriptive: In this approach, there was one single prompt for each question, to achieve this we removed the column descriptions if they couldn't be fitted in the context due to length limitations. The model's task was to predict the whole schema link generated previously. Using this approach some examples of the training set were omitted as the pure schema wouldn't fit in the context. 
    \item Chunked: In this approach, each question had N prompts, each with a maximum length of 4,096 tokens, as we were using the Llama-2 models. Both the schema and descriptions were included in each prompt preventing information loss. The model's task was to predict only the schema links present in the prompt and predict None if none of them were present.
\end{itemize}

\subsubsection{Not trusting hypothesis}

The training process begins by first training the schema linking model. After fine-tuning this model, we generated schema links for the entire data set. Subsequently, the model that generates the SQL is later trained using the schema link deduced by the first model and a resume version of the original schema. This approach offers the advantage of distributing the risk of inferring the structure of the database wrong to answer the question between the two models.

One notable limitation of this approach is that the schema link model generates better results for the training dataset, as the model has previously "seen" these examples, potentially biasing the SQL model to be more trustful of the schema link recommendations. However, due to the limited data, we decided to make this trade-off rather than further splitting the training dataset. 

\subsubsection{Trusting hypothesis}

Both the Schema-Link model and the SQL model are trained using the schema links constructed using reverse engineering from the BIRD dataset. This creates a trusting scenario where the SQL model trusts the output given by the schema link model, focusing solely on creating a syntactically and logically correct SQL query.

\section{Results}

We tested various models and combinations, the following section describes each one of them and the results obtained. We used the Dev partition of the BIRD-dataset and used the official \href{https://github.com/AlibabaResearch/DAMO-ConvAI/tree/main/bird}{Bird code} in order to evaluate our models.

\subsection{Vanilla}

First, we tested how the vanilla version of code-llama 13B would perform in the Bird-SQL dataset. In order to fit the schema and descriptions into the prompt we used a non-descriptive approach (${ND}_p$), that would leave out the column descriptions if they didn't fit in the prompt. We used 5000 tokens as the threshold.

In the prompt was specified that we only wanted the SQL query with no explanation but sometimes the model would also include an explanation of the query against what was prompted.

This led us to divide the vanilla model into two results "Vanilla" ($V$), which evaluates results exactly as the model outputs them, and "vanilla clean" ($V_c$) where we employed a post-processing step to extract only the SQL query from the generated results. 

Table 1 describes the execution accuracy results.

We conducted this test in order to understand the current capabilities of the models and set a floor value in order to improve upon it.

\newcolumntype{B}{>{\bfseries}c}
\begin{table}
  \caption{Results Vanilla models}
  \label{tab:freq}
  \begin{tabular}{ccccB}
    \toprule
    Model & Simple & Moderate & Challenging & Total\\
    \midrule
    ${ND}_p$ + $V$ & 20.76\% & 5.81\% & 1.39\% & 14.41\%\\
    \rowcolor[gray]{0.8}
    ${ND}_p$ + $V_c$ & 23.24\% & 7.96\% & 4.17\% & 16.82\%\\

  \bottomrule
\end{tabular}
\end{table}

\subsection{Zero shot fine-tune}

Later we fine-tuned a single code-llama 13B model in order to predict the SQL answer in a zero-shot environment (SQL$_{ft}$). In this model, we again used a non-descriptive approach (${ND}_p$) for the prompt as the only task required was to generate the final SQL query and this generation could not be divided into multiple prompts. The results are in the table 2. These results give us a comparison point between a one-step process and our approach of sub-dividing the problem into multiple steps.

\newcolumntype{B}{>{\bfseries}c}
\begin{table}
  \caption{Results direct fine-tune}
  \label{tab:freq}
  \begin{tabular}{ccccB}
    \toprule
    Model & Simple & Moderate & Challenging & Total\\
    \midrule
    ${ND}_p$ + $V$ & 20.76\% & 5.81\% & 1.39\% & 14.41\%\\
    ${ND}_p$ + $V_c$ & 23.24\% & 7.96\% & 4.17\% & 16.82\%\\
    \rowcolor[gray]{0.8}
    ${ND}_p$ + SQL$_{ft}$  & 35.79\% & 15.05\% & 5.56\% & 26.73\%\\

  \bottomrule
\end{tabular}
\end{table}

\subsection{Non-Trusting}

For our model combination, we first evaluated our non-trusting approach ($NT$). In this approach, the SQL model was trained with the schema links predicted by the Schema-Link model which were not always correct thus creating a non-trusting environment. This model was evaluated using a non-descriptive ($ND_p$) and a chunk approach ($CH_p$) for the schema link model. 

In the chunk approach ($CH_p$), we divide the schema with the column descriptions in N prompts, each prompt having a maximum threshold length. We generate predictions for each prompt and later join them to create the final schema link prediction.

Initially, we generated schema link predictions for both the non-descriptive and chunked schema link models. Subsequently, we provided these predictions to the SQL model, along with the schema and column descriptions incorporating the latter when they could be accommodated within the context without exceeding it.

\begin{table}
  \caption{Results Non-Trusting}
  \label{tab:freq}
  \begin{tabular}{ccccB}
    \toprule
    Model & Simple & Moderate & Challenging & Total\\
    \midrule
    ${ND}_p$ + $V$ & 20.76\% & 5.81\% & 1.39\% & 14.41\%\\
    ${ND}_p$ + $V_c$ & 23.24\% & 7.96\% & 4.17\% & 16.82\%\\
    ${ND}_p$ + SQL$_{ft}$  & 35.79\% & 15.05\% & 5.56\% & 26.73\%\\
    ${ND}_p$ + $SL$ + $NT$ & 42.16\% & 25.59\% & 19.44\% & 35.01\%\\
    \rowcolor[gray]{0.8}
    ${CH}_p$ + $SL$ + $NT$ & 52.97\% & 39.35\% & 29.86\% & 46.68\%\\
  \bottomrule
\end{tabular}
\end{table}

\subsection{Trusting}

In the "trusting" version of the model, we generated the schema link using the chunk approach immediately (${CH}_p$ + $SL$ + $T$). Subsequently, only the question and this prediction were prompted to the SQL model.

\begin{table}
  \caption{Results Trusting}
  \label{tab:freq}
  \begin{tabular}{ccccB}
    \toprule
    Model & Simple & Moderate & Challenging & Total\\
    \midrule
    ${ND}_p$ + $V$ & 20.76\% & 5.81\% & 1.39\% & 14.41\%\\
    ${ND}_p$ + $V_c$ & 23.24\% & 7.96\% & 4.17\% & 16.82\%\\
    ${ND}_p$ + SQL$_{ft}$  & 35.79\% & 15.05\% & 5.56\% & 26.73\%\\
    ${ND}_p$ + $SL$ + $NT$ & 42.16\% & 25.59\% & 19.44\% & 35.01\%\\
    \rowcolor[gray]{0.8}
    ${CH}_p$ + $SL$ + $NT$ & 52.97\% & 39.35\% & 29.86\% & 46.68\%\\
    ${CH}_p$ + $SL$ + $T$ & 32.22\% & 27.20\% & 9.03\% & 25.49\%\\
  \bottomrule
\end{tabular}
\end{table}

\subsection{Upper Bound}

Finally, we tested our model's upper bound, specifically the SQL model's upper bound, we did this in order to further understand our model's limitations. To test this we prompted the SQL model with the perfect schema links $SL_p$ extracted as detailed previously. This way the model would have the perfect recommendation of columns, tables and foreign keys to use. 

We also tested how the non-trusting model would perform if only the schema link and external knowledge were prompted in order to not confuse the LLM with extra content, the extra information was not necessary given that the schema links are perfect ($NT_{sl}$), Table 5 details our results.

\begin{table}
  \caption{Results Upper Bound}
  \label{tab:freq}
  \begin{tabular}{ccccB}
    \toprule
    Model & Simple & Moderate & Challenging & Total\\
    \midrule
    \rowcolor[gray]{0.8}
    $SL_p$ + $NT$ & 55.78  \% & 41.72\% & 29.86\% & 49.09\%\\
    $SL_p$ + $T$ & 53.73\% & 29.03\% & 24.31\% & 43.48\%\\
    $SL_p$ + $NT_{sl}$ & 35.46\% & 14.84\% & 11.81\% & 26.99\%\\
  \bottomrule
\end{tabular}
\end{table}

\subsection{Analysis}

In this section, we are going to contrast the results of the different models to see if our hypothesis were correct or not.

\subsubsection{Vanilla v/s Zero shot fine-tune}

The best vanilla results were 16.82\% ($ND_p + V_c$) and the worst of the fined-tuned models got 26.73\%. This shows an improvement of \textasciitilde10\% proving the hypothesis that trained models such as Llama get much better results when fine-tuned and are a powerful tool in resolving text-to-SQL problems.

\subsubsection{Trusting v/s Zero shot fine-tune}

The Trusting version did not perform well, as we can see in Table 4, getting only 25.49\% losing against the ${ND}_p$ + SQL$_{ft}$ version. 

This is due to the amount of responsibility that the schema linking step took. The input of the SQL model was solely the schema link, limiting the amount of context that the used model to create the SQL prediction. What happened in the majority of cases was that the $SL$ failed to capture a column needed or captured the name wrong, giving zero chance to the next step to get the answer well. From this, we learned two things:

\begin{enumerate}
    \item Getting the exact amount of information from the data schema that is needed to answer the question is not effective. It's better for the model to include extra columns than to risk having too few to make the right SQL prediction.
    \item Prompting the schema to the SQL model gives the chance to fix the $SL$ output, spreading the risk of not understanding the database architecture and enhancing the performance of the model greatly.
\end{enumerate}

\subsubsection{Non-Trusting v/s Zero shot fine-tune}

Comparing the results between our non-trusting approach and a zero-shot fine tune we can see a big spike in performance, providing evidence that dividing the tasks between models is beneficial when creating SQL queries. The improvement was \textasciitilde10\% for a non-descriptive setup (similar comparisons) and a  \textasciitilde20\% improvement for a chunked setup, providing further evidence that chunking the schema can lead to better results.

\subsubsection{Trusting v/s Non-Trusting Upper Bound}

To our surprise, the Non-trusting model performed better than the trusting one, this is a surprise because the trusting model was trained to create queries based solely on the schema links prompted to it, leaving it only the responsibility of creating SQL queries. A reason this could be the case is that after two epochs the trusting SQL model continued to learn i.e. the loss function on the validation set continually decreased and was stopped at the end of the second epoch.

To our surprise the model with just the schema linking ($NT_{sl}$) performed worse, this could be due to the fact that it was fine-tuned with a different prompt layout confusing the model more than helping it. One key takeaway is that fine-tuned models are very sensitive to prompt changes, further studies are required in order to explore this hypothesis.

\section{Conclusion}

Task division can be incredibly beneficial for LLM models, many agents and prompt chaining approaches have risen lately. In this study we proved how task division can be beneficial for NL2SQL tasks, increasing performance by more than 10\% compared to a direct approach. We also explored the idea and proposed a new framework to divide schema linking prompts in order to capture as much information as possible for the database schema and descriptions, this helped us improve by a further 10\% when compared to the non-descriptive approach. Finally, we tried combining two different models in order to further enhance the performance.
This study proves that smaller fine-tuned and task-specific approaches can deliver competitive results compared to GPT-4, it also presents an affordable way of fine-tuning models using easy access Google-colab notebooks. 

\section{Limitations}

There are a few significant limitations of this study, first of all, we couldn't test the impact of having 2 different models each specialized in one task, we encourage others to further validate the hypothesis that using LLama 2 for the schema link and Code-Llama for the SQL generation is actually beneficial.

The second limitation we encountered was coming up with a chunking method that would prevent information loss. Sometimes during our chunking two tables joined by a foreign key would get split up, although the table information containing the foreign key reference remained, the fact that the model didn't have access to the other table and its column could play an important part in picking the correct columns.

Finally, due to time limitations, we couldn't study what the new error composition looked like, this could lead to an interesting study of further modularizing the steps needed in order to generate a correct SQL.

\section{Acknowledgments}

Special thanks to Hans Löbel for his constant help and guidance during this process.

\bibliographystyle{ACM-Reference-Format}
\bibliography{sample-base}


\begin{thebibliography}{8}


\ifx \showCODEN    \undefined \def \showCODEN     #1{\unskip}     \fi
\ifx \showDOI      \undefined \def \showDOI       #1{#1}\fi
\ifx \showISBNx    \undefined \def \showISBNx     #1{\unskip}     \fi
\ifx \showISBNxiii \undefined \def \showISBNxiii  #1{\unskip}     \fi
\ifx \showISSN     \undefined \def \showISSN      #1{\unskip}     \fi
\ifx \showLCCN     \undefined \def \showLCCN      #1{\unskip}     \fi
\ifx \shownote     \undefined \def \shownote      #1{#1}          \fi
\ifx \showarticletitle \undefined \def \showarticletitle #1{#1}   \fi
\ifx \showURL      \undefined \def \showURL       {\relax}        \fi
\providecommand\bibfield[2]{#2}
\providecommand\bibinfo[2]{#2}
\providecommand\natexlab[1]{#1}
\providecommand\showeprint[2][]{arXiv:#2}

\bibitem[Gao et~al\mbox{.}(2023)]%
        {gao2023text}
\bibfield{author}{\bibinfo{person}{Dawei Gao}, \bibinfo{person}{Haibin Wang}, \bibinfo{person}{Yaliang Li}, \bibinfo{person}{Xiuyu Sun}, \bibinfo{person}{Yichen Qian}, \bibinfo{person}{Bolin Ding}, {and} \bibinfo{person}{Jingren Zhou}.} \bibinfo{year}{2023}\natexlab{}.
\newblock \showarticletitle{Text-to-SQL Empowered by Large Language Models: A Benchmark Evaluation}.
\newblock \bibinfo{journal}{\emph{arXiv preprint arXiv:2308.15363}} (\bibinfo{year}{2023}).
\newblock


\bibitem[Khot et~al\mbox{.}(2022)]%
        {khot2022decomposed}
\bibfield{author}{\bibinfo{person}{Tushar Khot}, \bibinfo{person}{Harsh Trivedi}, \bibinfo{person}{Matthew Finlayson}, \bibinfo{person}{Yao Fu}, \bibinfo{person}{Kyle Richardson}, \bibinfo{person}{Peter Clark}, {and} \bibinfo{person}{Ashish Sabharwal}.} \bibinfo{year}{2022}\natexlab{}.
\newblock \showarticletitle{Decomposed prompting: A modular approach for solving complex tasks}.
\newblock \bibinfo{journal}{\emph{arXiv preprint arXiv:2210.02406}} (\bibinfo{year}{2022}).
\newblock


\bibitem[Li et~al\mbox{.}(2023)]%
        {li2023can}
\bibfield{author}{\bibinfo{person}{Jinyang Li}, \bibinfo{person}{Binyuan Hui}, \bibinfo{person}{Ge Qu}, \bibinfo{person}{Binhua Li}, \bibinfo{person}{Jiaxi Yang}, \bibinfo{person}{Bowen Li}, \bibinfo{person}{Bailin Wang}, \bibinfo{person}{Bowen Qin}, \bibinfo{person}{Rongyu Cao}, \bibinfo{person}{Ruiying Geng}, {et~al\mbox{.}}} \bibinfo{year}{2023}\natexlab{}.
\newblock \showarticletitle{Can llm already serve as a database interface? a big bench for large-scale database grounded text-to-sqls}.
\newblock \bibinfo{journal}{\emph{arXiv preprint arXiv:2305.03111}} (\bibinfo{year}{2023}).
\newblock


\bibitem[Pourreza and Rafiei(2023)]%
        {pourreza2023din}
\bibfield{author}{\bibinfo{person}{Mohammadreza Pourreza} {and} \bibinfo{person}{Davood Rafiei}.} \bibinfo{year}{2023}\natexlab{}.
\newblock \showarticletitle{Din-sql: Decomposed in-context learning of text-to-sql with self-correction}.
\newblock \bibinfo{journal}{\emph{arXiv preprint arXiv:2304.11015}} (\bibinfo{year}{2023}).
\newblock


\bibitem[Rozi{\`e}re et~al\mbox{.}(2023)]%
        {roziere2023code}
\bibfield{author}{\bibinfo{person}{Baptiste Rozi{\`e}re}, \bibinfo{person}{Jonas Gehring}, \bibinfo{person}{Fabian Gloeckle}, \bibinfo{person}{Sten Sootla}, \bibinfo{person}{Itai Gat}, \bibinfo{person}{Xiaoqing~Ellen Tan}, \bibinfo{person}{Yossi Adi}, \bibinfo{person}{Jingyu Liu}, \bibinfo{person}{Tal Remez}, \bibinfo{person}{J{\'e}r{\'e}my Rapin}, {et~al\mbox{.}}} \bibinfo{year}{2023}\natexlab{}.
\newblock \showarticletitle{Code llama: Open foundation models for code}.
\newblock \bibinfo{journal}{\emph{arXiv preprint arXiv:2308.12950}} (\bibinfo{year}{2023}).
\newblock


\bibitem[Touvron et~al\mbox{.}(2023)]%
        {touvron2023llama}
\bibfield{author}{\bibinfo{person}{Hugo Touvron}, \bibinfo{person}{Louis Martin}, \bibinfo{person}{Kevin Stone}, \bibinfo{person}{Peter Albert}, \bibinfo{person}{Amjad Almahairi}, \bibinfo{person}{Yasmine Babaei}, \bibinfo{person}{Nikolay Bashlykov}, \bibinfo{person}{Soumya Batra}, \bibinfo{person}{Prajjwal Bhargava}, \bibinfo{person}{Shruti Bhosale}, {et~al\mbox{.}}} \bibinfo{year}{2023}\natexlab{}.
\newblock \showarticletitle{Llama 2: Open foundation and fine-tuned chat models}.
\newblock \bibinfo{journal}{\emph{arXiv preprint arXiv:2307.09288}} (\bibinfo{year}{2023}).
\newblock


\bibitem[Yu et~al\mbox{.}(2018)]%
        {yu2018spider}
\bibfield{author}{\bibinfo{person}{Tao Yu}, \bibinfo{person}{Rui Zhang}, \bibinfo{person}{Kai Yang}, \bibinfo{person}{Michihiro Yasunaga}, \bibinfo{person}{Dongxu Wang}, \bibinfo{person}{Zifan Li}, \bibinfo{person}{James Ma}, \bibinfo{person}{Irene Li}, \bibinfo{person}{Qingning Yao}, \bibinfo{person}{Shanelle Roman}, {et~al\mbox{.}}} \bibinfo{year}{2018}\natexlab{}.
\newblock \showarticletitle{Spider: A large-scale human-labeled dataset for complex and cross-domain semantic parsing and text-to-sql task}.
\newblock \bibinfo{journal}{\emph{arXiv preprint arXiv:1809.08887}} (\bibinfo{year}{2018}).
\newblock


\bibitem[Zhong et~al\mbox{.}(2017)]%
        {zhong2017seq2sql}
\bibfield{author}{\bibinfo{person}{Victor Zhong}, \bibinfo{person}{Caiming Xiong}, {and} \bibinfo{person}{Richard Socher}.} \bibinfo{year}{2017}\natexlab{}.
\newblock \showarticletitle{Seq2sql: Generating structured queries from natural language using reinforcement learning}.
\newblock \bibinfo{journal}{\emph{arXiv preprint arXiv:1709.00103}} (\bibinfo{year}{2017}).
\newblock


\end{thebibliography}

\appendix

\end{document}